\documentclass[conference]{ieeeconf}
\IEEEoverridecommandlockouts
\usepackage{cite}
\usepackage{amsmath,amssymb,amsfonts}
\usepackage{algorithmic}
\usepackage{graphicx}
\usepackage{textcomp}
\usepackage[table,xcdraw]{xcolor}
\usepackage[hidelinks]{hyperref}
\usepackage[T1]{fontenc}
\usepackage{color,soul}
\usepackage[font=small]{caption}
\usepackage{subcaption}
\usepackage{multicol}
\usepackage{multirow}
\usepackage{blindtext}
\usepackage{cuted}
\usepackage{adjustbox}
\usepackage{tabularx}
\usepackage{easytable}
\usepackage{float}
\usepackage{graphicx} 
\usepackage{hyperref}
\usepackage[linesnumbered,ruled,vlined]{algorithm2e}
\def\BibTeX{{\rm B\kern-.05em{\sc i\kern-.025em b}\kern-.08em
    T\kern-.1667em\lower.7ex\hbox{E}\kern-.125emX}}

\usepackage{siunitx} 
\usepackage{color, soul}

\begin{document}

\title{\bf Preference-Conditioned Reinforcement Learning for Space-Time Efficient Online 3D Bin Packing}
\author{
    Nikita Sarawgi$^{1}$,
    Omey M. Manyar$^{1}$,
    Fan Wang$^{2}$,
    Thinh H. Nguyen$^{2}$,
    Daniel Seita$^{1}$, 
    Satyandra K. Gupta$^{1}$
\thanks{$^{1}$ Viterbi School of Engineering, University of Southern California, Los Angeles, CA, USA. $^{2}$ Amazon Robotics, North Reading, MA, USA. Address all correspondence to \href{mailto:guptask@usc.edu}{guptask@usc.edu}}
}

\maketitle

\begin{abstract}
    Robotic bin packing is widely deployed in warehouse automation, with current systems achieving robust performance through heuristic and learning-based strategies. These systems must balance compact placement with rapid execution, where selecting alternative items or reorienting them can improve space utilization but introduce additional time. We propose a selection-based formulation that explicitly reasons over this trade-off: at each step, the robot evaluates multiple candidate actions, weighing expected packing benefit against estimated operational time. This enables time-aware strategies that selectively accept increased operational time when it yields meaningful spatial improvements. Our method, STEP (Space-Time Efficient Packing), uses a preference-conditioned, Transformer-based reinforcement learning policy, and allows generalization across candidate set sizes and integration with standard placement modules. It achieves a 44\% reduction in operational time without compromising packing density. Additional material is available at \href{https://step-packing.github.io}{https://step-packing.github.io}.
\end{abstract}

\section{Introduction}
\label{sec: Introduction}

Bin packing in industrial environments involves a sequence of tightly coupled operations, including reliable item picking, safe transport, and accurate placement within a bin. Packages exhibit diverse shapes, masses, and surface properties, which influence the feasibility and efficiency of robotic manipulation. 
Delays from failed attempts can quickly accumulate and degrade overall system performance, which is especially critical in large-scale automated warehouses. Consequently, time becomes a critical factor alongside space utilization in bin packing, requiring strategies that adapt to item-level variability to maintain overall throughput.

For efficient packing, bins must be filled compactly while delivering fast cycle times. 
Conventional methods that restrict grasps to the top face of an item~\cite{online3d_zhao_2021,zhao_PCT_2022, heng_GOPT_2024} achieve competitive packing density. 
However, such approaches do not account for cases where top-face grasps are unreliable or infeasible. Cuboidal items with varying dimensions can negatively influence future placements if not oriented appropriately. Recent methods permit 3D reorientation beyond simple in-plane rotations~\cite{Yin2024SolvingO3, Yin2025Learning3DBinPacking}, but they typically neglect the time overhead to reorient the robot's end-effector and to restore the item to a stable configuration for transport.

Human packers approach bin packing with an implicit trade-off between space efficiency and effort. Rather than always grasping from a fixed face, they choose whichever face allows them to pick, move, and place quickly, while ensuring efficient fits in the bin. This behavior suggests that efficient packing is not solely a spatial reasoning task, but also time-sensitive, involving appropriate action selection.


Inspired by this observation, we propose expanding the robot’s action space beyond top-face grasps to include alternative item choices and reorientations, thereby improving spatial efficiency and reducing packing failures. However, leveraging this flexibility requires accounting for the time overhead of picking, reorienting, and transporting of items and motivates strategies that must weigh time against spatial gains to maintain high throughput. In this paper, we demonstrate that our approach can achieve substantial reductions in operational time with marginal loss in packing efficiency, or alternatively, match the packing density of conventional methods while delivering faster cycle times.

\begin{figure}[t]
    \centering
    \includegraphics[width=\linewidth]{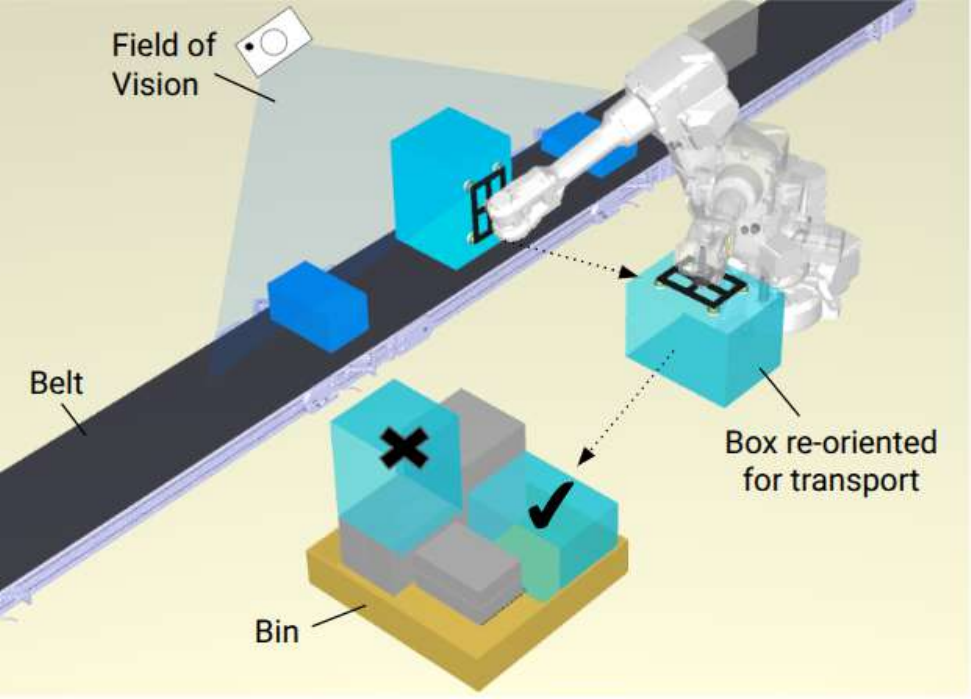}
    
    \caption{
        Visualization of the semi-online 3D bin packing problem (3D-BPP) with a buffer. The robot grasps the box from the front-face over the top-face. It reorients it to a transportable configuration and places it in the bin. The selected face impacts the placement position in the bin.
    }
    \label{fig:pull}
    \vspace{-19pt}
\end{figure}

To address this, we formulate the semi-online bin packing problem~\cite{Yin2025Learning3DBinPacking} as a multi-candidate selection problem that explicitly balances spatial utility and time-aware selections. 
This yields more favorable trade-offs along the Pareto front~\cite{hayes_MORL_2022}, achieving significant time savings while maintaining competitive packing quality. 

We propose \textbf{S}pace-\textbf{T}ime \textbf{E}fficient \textbf{P}acking (\textbf{STEP}), a Transformer-based~\cite{vaswani2017attention} multi-objective selection policy conditioned on user preferences, that captures dependencies among candidate items and the bin state. Self-attention models inter-item correlations, and cross-attention links item features to bin context, enabling joint reasoning over spatial and temporal factors to prioritize efficient selections.

In summary, our main contributions include:
\begin{itemize}
    \item A formulation of robotic bin packing as a multi-candidate selection problem over multiple items and their picking orientations, and explicit reasoning over the trade-off between spatial utility and time overhead
    \item A Transformer-based multi-objective selection policy that models spatial utility and operational cost of candidates through attention across items and bin context
    \item A modular and extensible framework that can be adapted to different robotic systems based on operational requirements
\end{itemize}



\section{Related Works}
\label{sec:related-works}

The Online 3D Bin Packing Problem (3D-BPP) has been studied extensively for decades \cite{christensen2017approximation, corcoran1992genetic}. Broadly, existing work can be categorized into two main methodological streams: (1) heuristic-based (classical) approaches \cite{li1992heuristic, monaci2006set}, and (2) learning-based (modern) approaches \cite{online3d_zhao_2021, wu2023machinelearningmultidimensionalbin, heng_GOPT_2024, zhang2021attend2packbinpackingdeep}, with some efforts attempting to combine the two \cite{yang2023heuristics}. Regardless of methodology, the central focus in almost all of these works has been to maximize space utilization. However, as discussed in Section~\ref{sec: Introduction}, the operational landscape in warehouse settings has shifted; thus, we also need to consider operational time for real-world adaptation. To clarify, by ``time,'' we do not refer to \emph{algorithmic} computation time, but rather the \emph{operational} time associated with performing physical actions, such as picking, and transporting, while maintaining the bin-packing efficiency. This operational time is influenced by several object-specific properties, such as shape, mass, surface characteristics, and graspability, and can significantly affect the overall efficiency of a bin packing system \cite{gompst_avigal_2022}. Yet, despite its practical importance, this dimension has received little attention in prior work.

Classical heuristic-based approaches, such as Next-Fit, First-Fit, and Harmonic families \cite{bays1977comparison}, have primarily relied on rule-based strategies that encode geometric constraints and bin state information, sometimes augmented with limited online buffers that provide access to more than one incoming item at a time \cite{zhang2017online}. While these heuristics are often computationally efficient and elegant, they rarely capture the realistic physical constraints that arise in deployment, especially those related to graspability, which directly contribute to operational time. Furthermore, the majority of these approaches either assume top-face picking by default or do not explicitly account for how different object orientations might affect process efficiency. Prior work on robotic package manipulation \cite{deformtransport_shukla_2024} has examined how such physical properties influence safe handling and placement feasibility, particularly for deformable items \cite{manyar2024simulation}, but with a focus on safety and success rates rather than throughput or time efficiency.

In recent years, learning-based methods have emerged as a popular alternative, where the online 3D-BPP is commonly framed as a Markov Decision Process and solved using deep reinforcement learning (DRL) \cite{online3d_zhao_2021, pmlr-v205-song23a, zhang2021attend2packbinpackingdeep}. Several works have attempted to hybridize learning and heuristics, e.g., using heuristics to guide the placement policy while learning a high-level item selection policy \cite{yang2023heuristics}. Other efforts, such as the Packing Configuration Tree (PCT) framework \cite{zhao_PCT_2022}, propose an explicit expansion of the combinatorial state-action space based on heuristics to improve search efficiency. While these approaches improve adaptability and can model complex state dependencies, they largely assume top-face picking and focus solely on spatial optimization. Recent work expands the action space to consider multiple grasping faces across available items \cite{Yin2025Learning3DBinPacking}, demonstrating that such orientation-aware selection can improve packing efficiency. However, these efforts treat orientation purely as a spatial decision variable and do not account for its impact on operational time. In contrast, our work explicitly incorporates time-sensitive handling costs into the selection process. By jointly reasoning over spatial utility and estimated operational time of different picking faces, our approach balances space efficiency with throughput, shifting the Pareto frontier to achieve substantial time savings while maintaining competitive packing efficiency.



\section{Problem Definition}

\subsection{Background}

Each of the stages of bin packing workflow (pick, transport, place) are subject to operational constraints that affect efficiency. Suction-based picks in logistics often fail due to surface and geometry~\cite{binpicking_2025_li}, leading to retries and added time.
Transport must respect object-specific motion constraints to ensure stability and safety~\cite{gompst_avigal_2022, criticalpp_pham_2019, deformtransport_shukla_2024}, while recovery from failures introduces further delays~\cite{recovery_matsuoka_2022}. To capture this complexity, we abstract these factors of pick-success, slower transport, or recovery overheads, into  a scalar time value to support time-aware decision making that adapts across robotic systems. Section~\ref{sec:real_experiments} illustrates how we formulate reorientation and transport costs for our setup.

\subsection{Problem Formulation}
We study the problem of semi-online 3D bin packing with item reorientation, where a robot sequentially packs a stream of rigid, cuboidal items with varying dimensions into a single bin, as illustrated in Figure~\ref{fig:pull}. Items arrive in unknown order, and only a partial set is available in the buffer at each step. 

We define the set of available items as \({\mathcal{O} = \{o_1, o_2, \ldots, o_N\}}\) of fixed size \(N\), where each item \(o_i \in \mathcal{O}\) is characterized by its dimensions \((l_i, w_i, h_i)\) in the on-conveyor camera frame. Each item exposes up to five graspable faces \(\mathcal{F} = \{\text{Top}, \text{Front}, \text{Back}, \text{Left}, \text{Right}\}\). To grasp a face \(f \in \mathcal{F}\), the robot must realign to its corresponding axis, incurring a reorientation time \(t_f\). Grasping and placing item $o_i$ from face $f$ incurs an additional transport and placement time $e_{i,f}$. The total operational time for item  \(o_i\) via face \(f\) is \({t_{i,f} = t_f + e_{i,f}}\), which we use throughout the paper as the time cost of packing the item through that face. Placement ensures that the grasped face becomes the top face in the bin, with added in-place rotations 
of \(90^\circ\) permitted about the axis orthogonal to the placement surface. The bin packing process follows standard physical constraints of static stability under gravity and inter-item contact and orthogonal placement aligned with the bin axes.

The bin is modeled as a 3D grid of size $(L, W, H)$, with the front-left-bottom corner at the origin and axes aligned with $X$, $Y$, $Z$ directions. Selection and placement update the bin state deterministically. Since we only study item selection, the placement positions of items in the bin are computed by an external placement function.

The objective is to maximize space utilization while minimizing cumulative operational time.
\textit{Space utilization} is  defined as the ratio of packed volume to bin volume:
\begin{equation}
    U = \frac{\sum_{i \in \mathcal{P}} V_i}{L \cdot W \cdot H}, \quad \text{with } V_i = l_i w_i h_i,
\label{eq:volume_reward}
\end{equation}
where \(\mathcal{P}\) is the set of items placed in the bin.

\textit{Operational time} captures the time cost of each placement:
\begin{equation}
    T = \sum_{i \in \mathcal{P}} (t_{i, f}),
\label{eq:time_reward}
\end{equation}
with $t_{i,f}$ the cost of placing item $o_i$ via face $f$.
The objective is therefore:
\begin{equation}
    \text{Optimize:} \quad
    \left\{
        \begin{array}{ll}
            \max \quad & U = \frac{\sum_{i \in \mathcal{P}} V_i}{L \cdot W \cdot H} \\
            \min \quad & T = \sum_{i \in \mathcal{P}} (t_{i, f})
        \end{array}
    \right.
\end{equation}

\section{Approach}

\begin{figure*}[t]
    \centering
    \includegraphics[width=\linewidth]{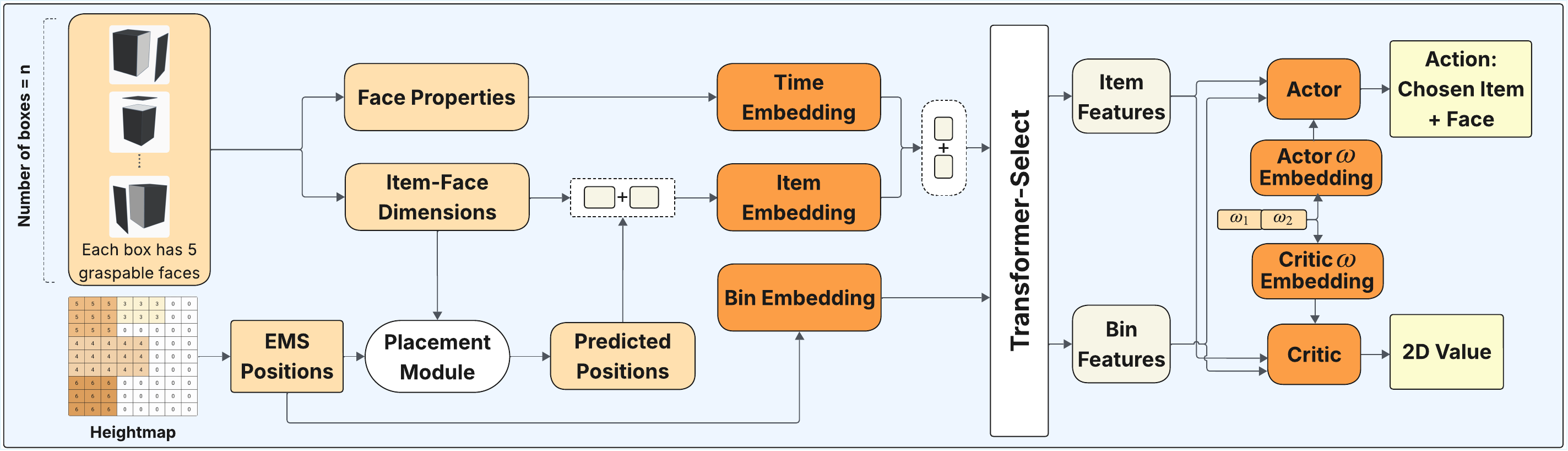}
    \vspace{-10pt}
    \caption{
    Overall architecture of \textbf{STEP}, a preference-conditioned Transformer-based policy network. The inputs consist of the bin state, item–face buffer with geometric and temporal attributes, and the current preference vector. These are embedded and processed by stacked attention-based encoders called Transformer-Select. The actor outputs logits over item–face candidates conditioned on the preference vector, while the preference-conditioned critic predicts vector-valued returns for space utilization and operational time, aligned with the multi-objective reward structure.
    }
    \label{fig:architecture}
    \vspace{-15pt}
\end{figure*}

\subsection{Reinforcement Learning Formulation}
We propose \textbf{STEP-n} (\textbf{S}pace–\textbf{T}ime \textbf{E}fficient \textbf{P}acking), where the suffix $n$ denotes the buffer size, i.e., the number of items available for selection at each decision step. To ensure that gains in space utilization do not come at the cost of increased operational time, the bin-packing problem is formulated as a Multi-Objective Markov Decision Process (MOMDP)~\cite{Sutton_2018} with dynamic preferences over weights assigned to each objective~\cite{momdppref_buet_2023}. Preferences are drawn from the space $\Omega$, where each $\omega \in \Omega$ satisfies $\sum_{i=1}^d \omega_i = 1$ and $\omega_i \geq 0$ for $d\ge1$ objectives. An MOMDP is defined by the tuple $\langle \mathcal{S_\omega}, \mathcal{A}, P_\omega, f_\omega(\textbf{r}), \gamma, \Omega \rangle$, where $\mathcal{S_\omega}$ denotes the state space augmented with the preference vector $\omega$ and $\mathcal{A}$ denotes the action space. The state transition function $P_\omega : \mathcal{S_\omega} \times \mathcal{A} \times \mathcal{S_\omega} \rightarrow [0, 1]$ defines the dynamics of the environment. The reward function $\textbf{r} : \mathcal{S_\omega} \times \mathcal{A} \times \mathcal{S_\omega} \rightarrow \mathbb{R}^d$ is vector-valued, with each component representing a distinct objective, such as spatial efficiency or operational time, that must be jointly optimized. The scalarization function $f_\omega : \textbf{r} \rightarrow \mathbb{R}$ maps the vector reward to a scalar under preference $\omega$. The discount \(\gamma \in [0, 1)\) weights future rewards.

\subsubsection{\textbf{Preference Weight}}
Each preference vector \(\omega \in \Omega\) is a two-dimensional vector \([\omega_1, \omega_2]\) where \(\omega_1\) and \(\omega_2\) denote the weights for spatial efficiency and operational time, respectively. 
Both \(\omega_1\) and \(\omega_2\) are nonnegative, and \(\omega_1 + \omega_2 = 1\). The preference space \(\Omega\) is constructed by uniformly sampling 50 such vectors from this interval, resulting in a fixed discrete distribution of trade-offs between the two objectives.

\subsubsection{\textbf{State}} 
The state $s_t \in \mathcal{S_\omega}$ at each time step $t$ consists of four parts: the bin configuration $s_{t,\text{bin}}$, available item state $s_{t,\text{buffer}}$, the time state \(s_{t,\text{time}}\), and the current preference vector $\omega_t$. Following prior work~\cite{heng_GOPT_2024,zhao_PCT_2022}, the bin state $s_{t,\text{bin}}$ is represented as a fixed-length sequence of $N_{\text{EMS}}$ Empty Maximal Spaces (EMS)~\cite{ems_parreno_2008}, where each EMS encodes a free volume currently available for placement. If fewer than $N_{\text{EMS}}$ EMS are available, the sequence is zero-padded.
Each EMS is encoded as a 6D vector containing the coordinates of its front-left-bottom and top-right-back corners in 3D space. For the buffer state \(s_{t,\text{buffer}}\), following the formulation in~\cite{Yin2025Learning3DBinPacking}, we treat each graspable face of an item as an independent decision unit. This encourages the policy to evaluate each placement option independently. The top face is represented by the item dimensions as observed from the conveyor’s camera frame, $(l_i, w_i, h_i)$; the front and back faces are aligned as \((l_i, h_i, w_i)\); and the left and right faces are represented as \((w_i, h_i, l_i)\).
For a buffer of $N$ items, each with five graspable faces, the state is represented by $5N$ item–face pairs.
Each unit is encoded as a 7-dimensional vector: (i) the item’s effective dimensions when grasped from face $f$; (ii) the predicted front-left-bottom (FLB) position $(x_{i,f}, y_{i,f}, z_{i,f})$ for stable placement in the bin; and (iii) a binary rotation flag $r_{i,f} \in \{0,1\}$ indicating whether a $90^\circ$ in-plane rotation is applied. The time state \(s_{t,\text{time}}\) encodes the scalar operational time \(t_{i,f}\) for each item-face unit.

The placement positions are predicted by an external module, which may be either heuristic or learned. We use GOPT~\cite{heng_GOPT_2024}, a competitive learning-based policy for computing placements in 3D bin packing, which takes the effective dimensions of an item-face unit and returns the front-left-bottom (FLB) coordinates and its rotation flag.



Empirical evaluations on our robotic setup showed operational time varies with grasp direction and surface characteristics. Front-face reorientation is fastest, side faces incur moderate delay, and back faces are slowest. We model this using fixed reorientation penalties: 1 (front), 2 (left/right), and 3 (back). Transport time after a successful grasp also depends on the surface type due to suction-cup instability. To capture this, each face is randomly assigned one of three categories: smooth (0), taped/plastic-covered (2), or packaging-labeled (4). The corresponding penalty is added to the operational time of the item-face pair. In deployment, these values could be derived from grasp-conditioned trajectory times, but here we represent them as scalar costs to focus on the selection-level trade-off.

\subsubsection{\textbf{Action}}
The action at time $t$, denoted $a_t \in \mathcal{A}$, is to select an index from the item candidate set, which specifies both the item and its associated graspable face.  Each of the $N$ candidate items exposes five graspable faces, resulting in an action space of $5 \times N$ discrete choices.

\subsubsection{\textbf{State Transition}} 
At time $t$, the selected item is placed at its predicted location. If the rotation flag is set, the item is rotated by $90^\circ$ around the axis perpendicular to the chosen face, which defines the grasping orientation and becomes the top face after placement. The EMS set updates to reflect the occupied volume, and a new item is added to the buffer.

\subsubsection{\textbf{Reward}}  
The agent receives a two-dimensional vector:
\[
\mathbf{r}_t = \big[ r^{\text{space}}_t, \; r^{\text{time}}_t \big],
\]
where $r^{\text{space}}_t$ represents the volume gain from the placed item (Equation~\ref{eq:volume_reward}), and $r^{\text{time}}_t$ denotes the operational cost for grasping, transporting and placing it (Equation~\ref{eq:time_reward}).

\subsubsection{\textbf{Scalarization Function $f_\omega$}}
We define a linear scalarization function $f_\omega$ that maps the two-dimensional reward vector $\mathbf{r}_t = [r^{\text{space}}_t, r^{\text{time}}_t]$ to a scalar value using the current preference vector $\omega_t = [\omega_{t,1}, \omega_{t, 2}] \in \Omega$. The scalarization is defined as:
$
f_\omega(\mathbf{r}_t) = \omega_t^\top \mathbf{r}_t,
$
which in this case evaluates to:
\[
f_\omega(\mathbf{r}_t) = \omega_{t,1} \cdot r^{\text{space}}_t + \omega_{t,2} \cdot r^{\text{time}}_t.
\]
This formulation allows the agent to reason over trade-offs between spatial efficiency and operational time based on the current preference weighting.

\subsection{Network Architecture}

\begin{figure}[t]
    \centering
    \includegraphics[width=0.6\linewidth, height=5cm]{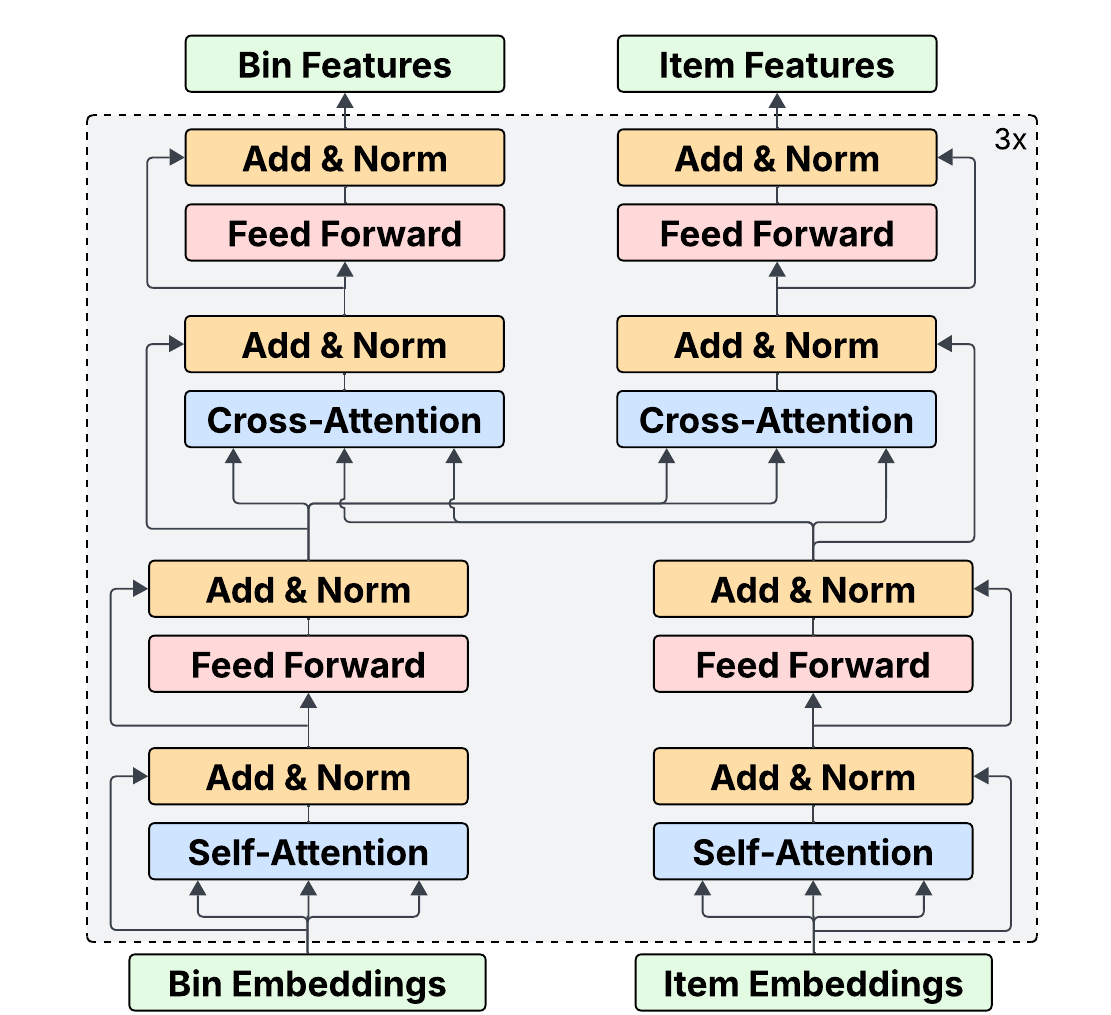}
    \caption{
        The Transformer-Select module used in our proposed method \textbf{STEP}. 
    }
    \label{fig:method}
    \vspace{-18pt}
\end{figure}

As in prior bin-packing work~\cite{heng_GOPT_2024, Yin2025Learning3DBinPacking}, we use a Transformer encoder~\cite{vaswani2017attention} to model correlations between items and bin features. The network architecture consists of four components: input embedding layers, an attention-based encoder stack, named Transformer-Select, and separate output heads for the actor and critic.
As shown in Figure~\ref{fig:architecture}, it takes three input components: the bin representation \(s_{t,\text{bin}}\) $\in$ $\mathbb{R}^{N_{\text{EMS}} \times 6}$, the buffer state \(s_{t,\text{buffer}}\) $\in$ $\mathbb{R}^{5N \times 7}$ and the time state \(s_{t,\text{time}}\) $\in$ $\mathbb{R}^{5N \times 1}$.
These inputs are individually processed by Multi-Layer Perceptrons (MLPs), after which the item and time embeddings are concatenated to preserve temporal cost as an explicit signal and prevent it from being absorbed into geometric features during attention.

The Transformer-Select (Figure~\ref{fig:method}) is constructed of three stacked encoder blocks, each containing two self-attention layers and a bi-directional cross-attention module. Self-attention is applied independently to EMS and item-face embeddings to capture intra-set structure. The cross-attention module includes EMS-to-item and item-to-EMS layers, enabling the model to reason jointly over bin configuration and item candidates. Each attention layer is followed by a feedforward MLPs and layer normalizations.

The actor and critic networks are implemented as separate MLPs. The bin and item features from the Transformer-Select are individually conditioned on the learned embedding of the current preference vector $\omega_t$. The actor computes scaled dot-product similarities between the embeddings, which are averaged across EMS tokens and passed through a learned bias to produce logits over the selection candidates. The critic concatenates the embeddings and passes the result through an MLP that outputs a two-dimensional vector, where each component estimates the expected discounted return for spatial efficiency and operational time. This vectorized value function aligns with the structure of the multi-objective reward in the MOMDP formulation.

\begin{figure*}[t]
    \centering
    \includegraphics[width=0.9\linewidth, height=5.5cm]{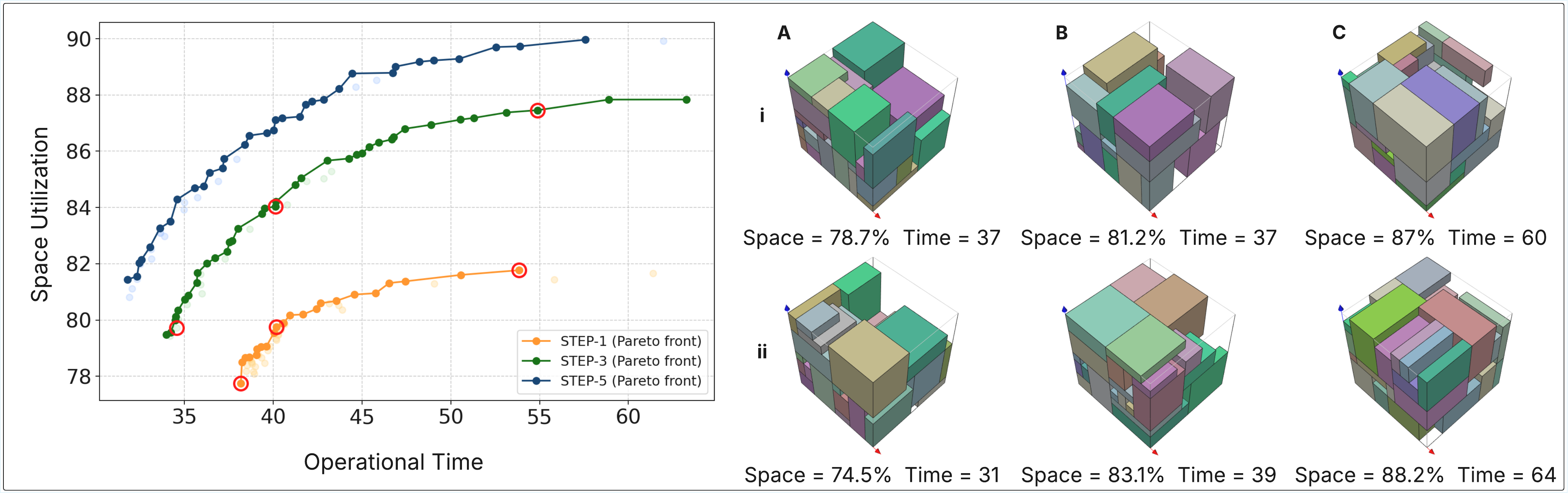}
    \caption{
       Pareto front of evaluated policies in the space–time trade-off space (left). Each point represents a distinct policy configuration, with the frontier highlighting non-dominated solutions balancing space utilization and operational time. Packing visualizations for STEP-1 (i) and STEP-3 (ii) at three preference vectors A = $[0.02, 0.98]$, B = $[0.53, 0.47]$, and C = $[0.95, 0.05]$ are shown (right), where the first element weights space and the second weights time. The corresponding $\omega$ values are marked with red circles in the Pareto front.
}
    \label{fig:pareto_packing_visualization}
    \vspace{-15pt}
\end{figure*} 

\subsection{Training Method}

We adopt the Robust Dynamic Preferences Multi-Objective Reinforcement Learning (RDP-MORL) framework~\cite{momdppref_buet_2023} to train a single preference-conditioned policy capable of balancing spatial efficiency and operational time. This approach has shown effectiveness in domains such as internet network management~\cite{heo2024dynamicmorl}, motivating its application to our method where dynamic trade-offs must be managed. We integrate RDP-MORL into the Proximal Policy Optimization (PPO) algorithm to handle vector-valued rewards under dynamic preferences.


The critic estimates a vector-valued value function \(\mathbf{V}_\pi(s) \in \mathbb{R}^2\), defined as:
\begin{equation}
\mathbf{V}_\pi(s) = \mathbb{E}_\pi \left[ \sum_{i=0}^{\infty} \gamma^i \mathbf{r}_{t+i+1} \;\middle|\; s_t = s \right],
\end{equation}
where \(\gamma \in [0,1)\) is the discount, and \(\mathbf{r}_{t} \in \mathbb{R}^2\) is the vector reward at time \(t\). It is trained by minimizing the squared error between the predicted and empirical vector returns:
\begin{equation}
\mathcal{L}_{\text{critic}} = \left\| \hat{\mathbf{V}}(s_t, \omega) - \sum_{i=0}^{T-t} \gamma^i \mathbf{r}_{t+i} \right\|_2^2.
\end{equation}

We use Generalized Advantage Estimation (GAE)~\cite{schulman2016GAE} to compute a vector of advantages \(\mathbf{A}_t \in \mathbb{R}^2\), which are then scalarized using the sampled preference vector \(\omega\), such that $\hat{A}_t^\omega = \omega^\top \mathbf{A}_t$. The scalarized advantage \(\hat{A}_t^\omega\) is used to optimize the actor via the PPO clipped objective:
\begin{equation}
\mathcal{L}_{\text{clip}} = \mathbb{E}_t \left[ \min \left( \rho_t \hat{A}_t^\omega,\; \text{clip}(\rho_t, 1 - \epsilon, 1 + \epsilon) \hat{A}_t^\omega \right) \right],
\end{equation}
\vspace{-3pt}
where \(\rho_t\) is the policy ratio and \(\epsilon\) is the PPO clipping threshold.
This framework allows the agent to learn a single policy that dynamically adapts to varying objective preferences by conditioning on \(\omega\), enabling robust performance across different trade-offs between space and time.

\vspace{-4pt}
\section{Experiments}

\subsection{Implementation Details}

Our method is implemented in PyTorch by adapting the multi-objective PPO algorithm, MOPPO, from the \texttt{morl-baselines} library~\cite{felten_morl_2023} into the Tianshou reinforcement learning framework~\cite{tianshou_weng_2022}. A preference vector is uniformly sampled from a set of 50 predefined weights at the start of each episode. The training batch size is 128 and the step number for GAE updates is 3, with \(\lambda_{\text{GAE}} = 0.97\) and discount factor \(\gamma = 1\). We use the Adam optimizer with a learning rate of \(3 \times 10^{-5}\). For PPO loss, the value and entropy loss coefficients are set to \(c_1 = 0.5\) and \(c_2 = 0.001\), with a clipping ratio \(\epsilon = 0.3\).  We use the RS dataset~\cite{online3d_zhao_2021} for training and evaluation. Bin dimensions are fixed at \(10 \times 10 \times 10\). To introduce variation in box sizes, item dimensions \(l_t, w_t, h_t\) are sampled as \( \frac{\min(L, W, H)}{10} \leq l_t, w_t, h_t \leq \frac{\min(L, W, H)}{2}\).

\renewcommand{\arraystretch}{1.3} 
\begin{table*}[!ht]
\centering
\caption{Comparison of bin packing performance across space and time metrics for different methods.}
\label{tab:comparison}
\begin{tabular}{>{\hspace{4pt}}l<{\hspace{4pt}}>{\hspace{4pt}}c<{\hspace{4pt}}>{\hspace{4pt}}c<{\hspace{4pt}}>{\hspace{4pt}}c<{\hspace{4pt}}>{\hspace{4pt}}c<{\hspace{4pt}}>{\hspace{4pt}}c<{\hspace{4pt}}>{\hspace{4pt}}c<{\hspace{4pt}}>{\hspace{4pt}}c<{\hspace{4pt}}>{\hspace{4pt}}c<{\hspace{4pt}}>{\hspace{4pt}}c<{\hspace{4pt}}}
\hline
\textbf{Method} & \textbf{Uti} & \textbf{Var} & \textbf{Num} & \textbf{Time}  & \textbf{VarTime}  & \textbf{Top \(\%\)}  & \textbf{Front \(\%\)}  & \textbf{Left+Right \(\%\)}  & \textbf{Back \(\%\)}\\
\hline
TopFaceSpace & 75.59  & 7.25 & 28.95 & 57.87 & 14.31 & 100.00 & 0.00 & 0.00 & 0.00 \\
ReorientSpace-1 & 84.05 & 5.64 & 32.26 & 96.60 & 20.01 & 32.87 & 34.45 & 32.67 & 0 \\
\rowcolor[HTML]{E0FFC9}
ReorientTime-1 & 75.98 & 6.97 & 29.07 & 36.08 & 9.30  & 59.07 & 27.04 & 12.16 & 1.73 \\
\rowcolor[HTML]{C2FF92}
STEP-1 (Ours)  & 81.76 & 6.16 & 31.55 & 53.84 & 12.46 & 49.79 & 26.50 & 20.77 & 2.92 \\
\hline
\end{tabular}
\vspace{-12pt}
\end{table*}



\subsection{Exploring the Space-Time Trade-off}
To examine the trade-offs in our bin packing formulation, we plot the Pareto front of the learned convex coverage set in Figure~\ref{fig:pareto_packing_visualization}, following~\cite{hayes_MORL_2022}. Each point shows the space utilization and operational time achieved by the same preference-conditioned policy under a unique preference vector $\omega \in \Omega$, sampled uniformly from the 2D simplex over $[0,1]^2$. The points on the Pareto front correspond to non-dominated outcomes, i.e., there is no other policy that improves the expected return for one objective without reducing it for another. We compare the Pareto fronts obtained from the policy tested with buffer sizes of 1, 3, and 5 candidate items. 

The plot exhibits a consistent trend where higher space utilization corresponds to increased operational time, reflecting the inherent trade-off between the two objectives. It illustrates how weighted-objective selection enables controllable trade-offs between space utilization and operational time. When additional time is permissible, the policy favors tighter placements to achieve higher packing efficiency. Conversely, under stricter time budgets, it prioritizes faster execution at the expense of space utilization. As expected, increasing the buffer size leads to improved space utilization due to greater selection flexibility. The trade-off between space utilization and time cost remains consistent across buffer sizes. This performance curve enables practitioners to select operating points aligned with application-specific constraints.


\subsection{Performance Evaluation of Face Selection}
We evaluate the impact of face selection on space and time by comparing policy variants with different orientation strategies. To isolate this effect, all variants are tested in a single-item buffer where only one item is available per timestep.
As our focus is on analyzing trade-offs in space and time with single items, we consider GOPT~\cite{heng_GOPT_2024} configured to select the top face of each item, allowing \(90^\circ\) in-plane rotations. We refer to it as \textbf{TopFaceSpace} in our experiments.
\textbf{ReorientSpace-1} enables selection among three symmetric grasp faces of a cuboidal item, with \(90^\circ\) in-plane rotations allowed for each. This setup is equivalent to the single-item buffer case in~\cite{Yin2025Learning3DBinPacking}, which considers six possible orientations and is trained solely to maximize space utilization without modeling time costs.
We also introduce \textbf{ReorientTime-1} as a time-optimization baseline, implemented to enable face selection while optimizing exclusively for operational time and disregarding spatial efficiency in its selection.
We evaluate our proposed method, denoted \textbf{STEP-1}, at the preference vector \(\omega = (0.95, 0.05)\), which yields the highest space utilization among all configurations.
In all method names, the numerical suffix indicates the buffer size.
All these variants share GOPT~\cite{heng_GOPT_2024} as their placement module, and differ only in their item's face selection strategy. 
See Table~\ref{tab:comparison} for quantitative results.

We evaluate performance across five metrics: average space utilization (\textit{Uti}) and its standard deviation (\textit{Var}); average number of successfully packed items(\textit{Num}); average time to reach a terminal configuration (\textit{Time}), and its standard deviation (\textit{VarTime}). We report face selection as the proportion of grasps from five directions: \textit{Top}, \textit{Front}, \textit{Back}, \textit{Left}, and \textit{Right}. Since \textit{Left} and \textit{Right} incur similar reorientation costs, they are aggregated into one category in Table~\ref{tab:comparison}.

Since each face incurs operational cost ($t_{i, f}$), TopFaceSpace reports non-zero time even with top-face grasps without reorientation. A comparison of TopFaceSpace and ReorientSpace-1 highlights the impact of face selection in packing performance. Both use a single-item buffer, but ReorientSpace-1 selects uniformly from three graspable orientations, yielding an \textbf{$8.46\%$} gain in space utilization, attributable to reorientation flexibility, at the cost of higher handling time. ReorientTime-1 and ReorientSpace-1 further contrast objective priorities: ReorientTime-1 minimizes operational time by favoring quick grasps and achieves packing density comparable to TopFaceSpace. \textbf{STEP-1} achieves $81.76\%$ space utilization, $6.17\%$ higher than top-face-only selection, and also $5.62\%$ higher than the time-optimized baseline ReorientTime-1. 
While ReorientSpace-1 achieves the highest space utilization, it also incurs the highest operational time. In contrast, STEP-1 achieves comparable space efficiency, with only $2.29\%$ loss, while reducing operational time by $44\%$. This key result highlights that jointly optimizing for space and time enables competitive space efficiency without sacrificing throughput.

\subsection{Generalization on Buffer Size}
We evaluate the policy with buffers of 1, 3, and 5 items, despite training only with a buffer of size 5.
Using the same preference vector $\omega = (0.95, 0.05)$ as before, we assess adaptation under both space and time focused objectives. Results in Table~\ref{tab:general_buffer_table} show space utilization increases by $7.96\%$ as the buffer size grows from 1 to 5. Operational time remains nearly constant, demonstrating that the policy exploits larger candidate sets to achieve denser packing without incurring extra time cost. Compared to single-item top-face baseline (TopFaceSpace), STEP-5 shows $14.13\%$ increase in space utilization while still keeping the operational time low. This is a key result highlighting the efficiency of the approach.

\begin{table}[H]
\small
\centering
\caption{Generalization performance of our proposed method across different buffer sizes, evaluated on space utilization and operational time using a fixed preference vector $\omega = (0.95, 0.05)$. The suffix indicates the buffer size.}
\label{tab:general_buffer_table}
\begin{tabular}{lccccc}
\hline
\textbf{Method} & \textbf{Uti} & \textbf{Var} & \textbf{Num} & \textbf{Time} & \textbf{VarTime} \\
\hline
STEP-1  & 81.76 & 6.16 & 31.55 & 53.84 & 12.46 \\
STEP-3 & 87.45 & 3.82 & 34.10 & 54.89  & 12.01 \\
STEP-5  & 89.72 & 3.21 & 34.96  & 53.89  & 11.68  \\
\hline
\end{tabular}
\vspace{-8pt}
\end{table}

\subsection{Effect of Variability in Items}
We evaluate the impact of item geometry on packing performance, measured in terms of space utilization and operational time. Items are divided into two geometric categories: \textit{uniform} and \textit{variable}. An item is \textit{uniform} if its three dimensions are within two units of each other, yielding a near-cubic geometry. Items are labeled \textit{variable} when at least two dimensions differ by three or more units, producing elongated, tall, or flat shapes. We evaluate performance on mixed datasets with controlled item variability, using items with dimensions sampled from the range $[1, 6]$ within a $10 \times 10 \times 10$ bin.

\begin{figure}[t]
    \centering
    \includegraphics[width=\linewidth]{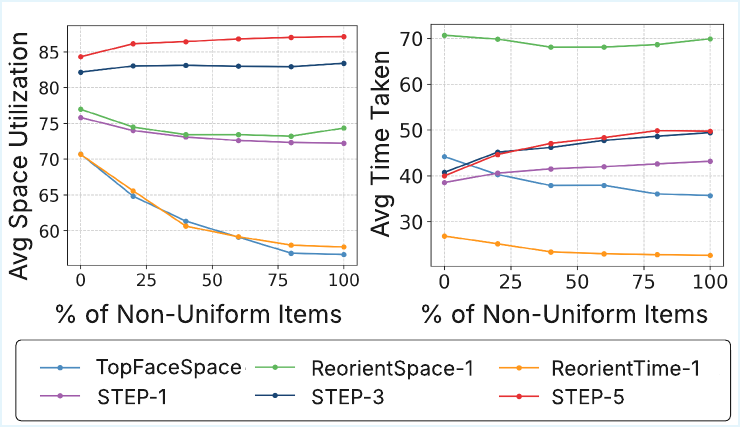}
    \caption{
        {Average Space Utilization ($\%$) and Average Time Taken under increasing variability in item box sizes.}
    }
    \label{fig:variability_space_time_plot}
    \vspace{-15pt}
\end{figure}

Figure~\ref{fig:variability_space_time_plot} shows average space utilization and average operational time as a function of the percentage of variable shaped items for six policies: three baselines (TopFaceSpace, ReorientSpace-1, ReorientTime-1) and our proposed methods (STEP-1, STEP-3, STEP-5). As item variability increases, TopFaceSpace, which is restricted to top-face grasps, exhibits a sharp decline in space efficiency. In contrast, ReorientSpace-1 and the STEP policies maintain stable space efficiency, with only minor fluctuations, due to their ability to exploit reorientation during placement. ReorientTime-1, which ignores space in decision-making, mirrors TopFaceSpace trends for space efficiency since it uses GOPT for placement decisions.
Access to non-top faces is critical for variable-sized items, where alternative orientations allow tighter, more stable fits. Near-cubic items fit well without reorientation, explaining higher utilization under policies that ignore it.

TopFaceSpace shows reduced operational time with increasing item variability, primarily since its inability to reorient leads to uneven bin filling, and early termination with unplaceable items. ReorientTime-1 remains largely unaffected by shape variability in terms of operational time, as its selection strategy consistently prioritizes grasps with minimal time. ReorientSpace-1 yields the highest operational time across all variability levels. In contrast, STEP-1 incurs slight increase in time with higher item variability, but maintains space efficiency. It learns to select faces and items that enable tighter packing without excessive operational costs. Importantly, increasing the buffer size in STEP-3 and STEP-5 show the highest space efficiency, with a gradually increasing time cost as item variability increases.

\begin{table}[H]
\small
\centering
\caption{Comparison of space and time performance across MCTS, GOPT variants, ReorientSpace, and STEP-5.}
\label{tab:mcts_comparison}
\begin{tabular}{lccccc}
\hline
\textbf{Method} & \textbf{Uti} & \textbf{Var} & \textbf{Num} & \textbf{Time} & \textbf{VarTime} \\
\hline
GOPT\_var1 & 83.01 & 6.72 & 31.78 & 95.59 & 19.80 \\
ReorientSpace-1 & 84.05 & 5.64 & 32.26 & 96.60 & 20.01 \\
\cline{1-6}
MCTS-5 & 54.58 & 18.55 & 13.41 & 46.95 & 19.21 \\
ReorientSpace-5 & 91.26 & 3.35 & 35.71 & 106.71 & 20.12 \\
STEP-5 & 89.72 & 3.21 & 34.96  & 53.89  & 11.68 \\
\hline
\end{tabular}
\vspace{-10pt}
\end{table}

\subsection{Comparison with Selection Baselines}

We evaluate our proposed method \textbf{STEP-5}, which operates with a buffer of five items, against two baselines. The first, ReorientSpace-5, is trained to reorient items across six orientations (3 graspable faces with 2 in-plane rotations each) for a buffer of five items, following the setup in~\cite{Yin2025Learning3DBinPacking}. The second baseline is based on Monte Carlo Tree Search (MCTS), a strategy previously used for buffer-based item selection in bin packing~\cite{online3d_zhao_2021, zhao_PCT_2022}. Our MCTS implementation uses 50 simulations per decision step, with the search tree restricted to item–face actions. Item selection is guided by a reward function defined as a linear scalarization of space utilization and operational time, with preference weights matching those of STEP. All results are averaged over 100 evaluation episodes.

To isolate the effect of learning multi-objective item selection, we extend the GOPT architecture~\cite{heng_GOPT_2024}, originally designed for placement, to predict both the item and grasp face. We test three variants. GOPT\_var1 selects among three principal faces of a single item with \(90^\circ\) in-plane rotations (six orientations), trained only for space utilization without time awareness. Its performance is compared against ReorientSpace-1. The second variant extends to multiple buffer items, each with three graspable faces and in-plane rotations. The third adds time-awareness, using a scalarized reward for joint optimization of space and time, mirroring our main objective. Both extensions showed unstable training, with inconsistent rewards and failure to converge, highlighting the importance of decoupling item selection from placement as in our approach for multiple objectives.

Quantitative results are shown in Table~\ref{tab:mcts_comparison}. GOPT\_var1 performs comparably to ReorientSpace-1 in both space utilization and time, as both are optimized only for spatial efficiency in single-item settings. However, it fails to scale to multi-item buffers, which ReorientSpace handles effectively.
Our method, STEP-5, outperforms MCTS-5 in both space utilization and number of items packed. Although MCTS-5 reports lower operational time, this is primarily due to fewer items being packed per episode, and it suffers from high computational overhead caused by tree expansion. 
Similar to the single-item setting, ReorientSpace-5 achieves the highest space utilization but with very high operational time. STEP-5 reduces time by 49\% at the cost of only 1.54\% in space efficiency, demonstrating the benefit of time-aware selection.


\begin{figure}[htbp]
    \centering
    
    \begin{subfigure}{0.59\columnwidth}
        \centering
        \includegraphics[width=\linewidth]{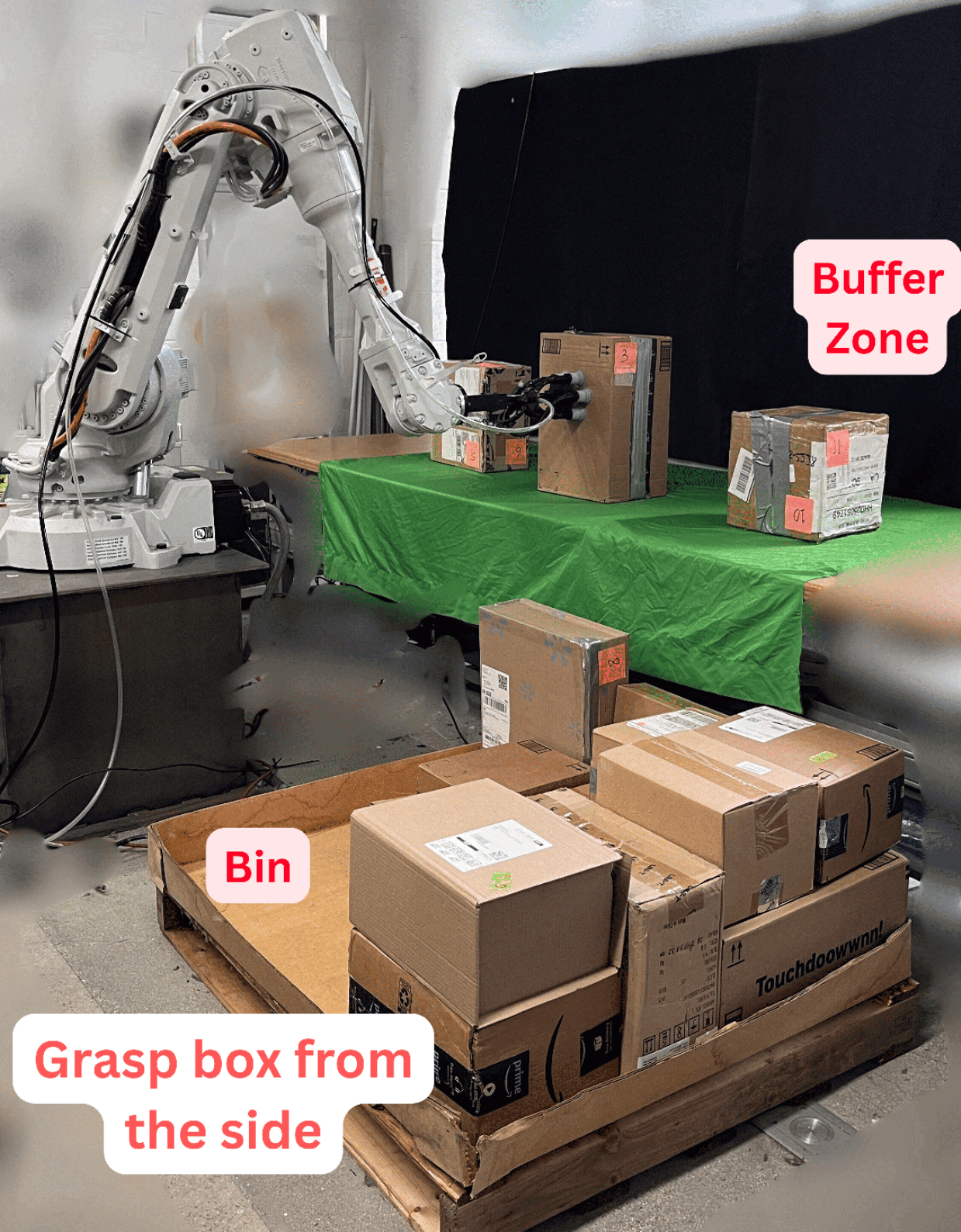}
        \vspace{-15pt}
        \caption{}
    \end{subfigure}
    \begin{subfigure}{0.39\columnwidth}
        \centering
        \includegraphics[width=\linewidth]{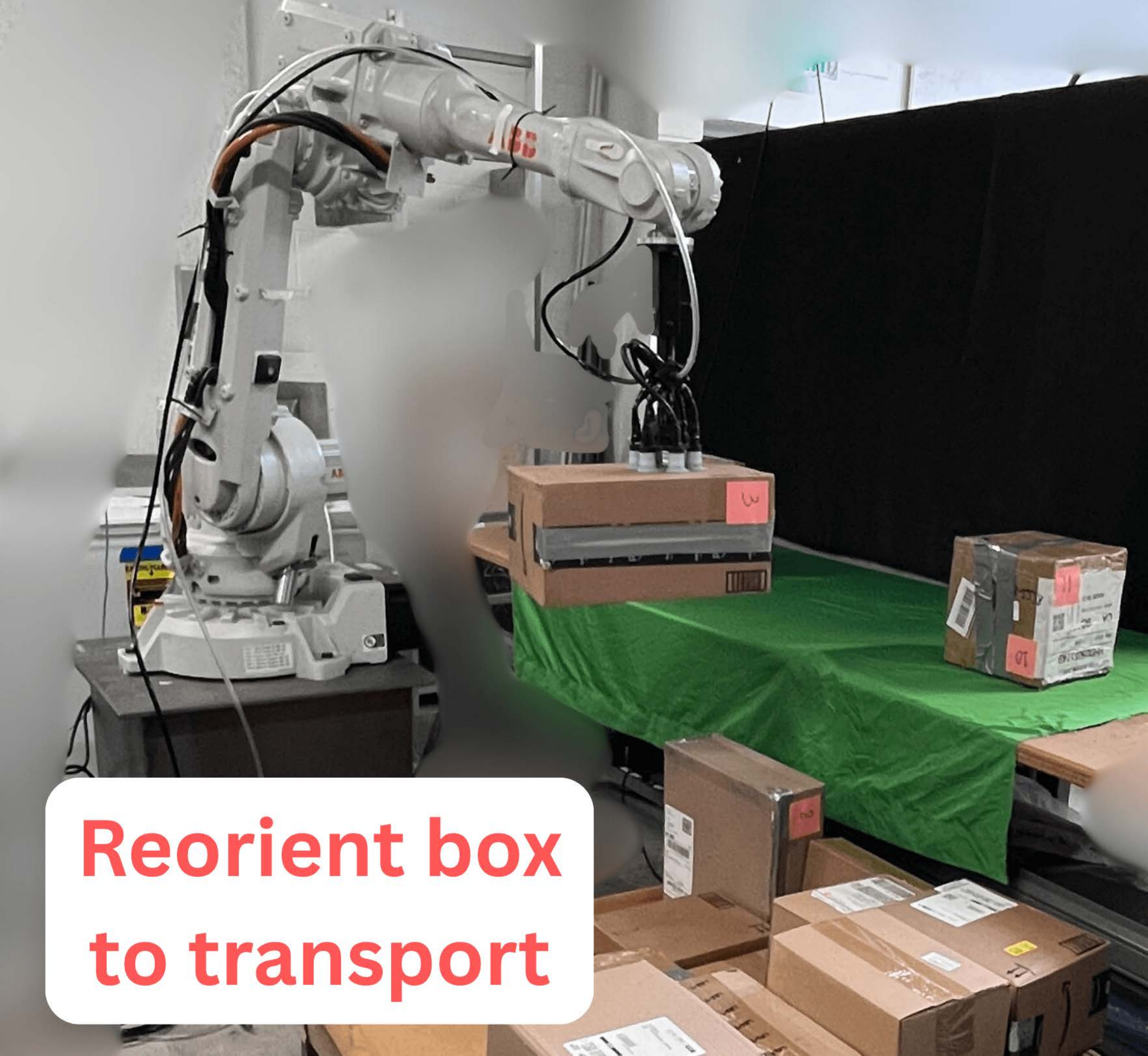}
        \vspace{-15pt}
        \caption{}
        
        \includegraphics[width=\linewidth]{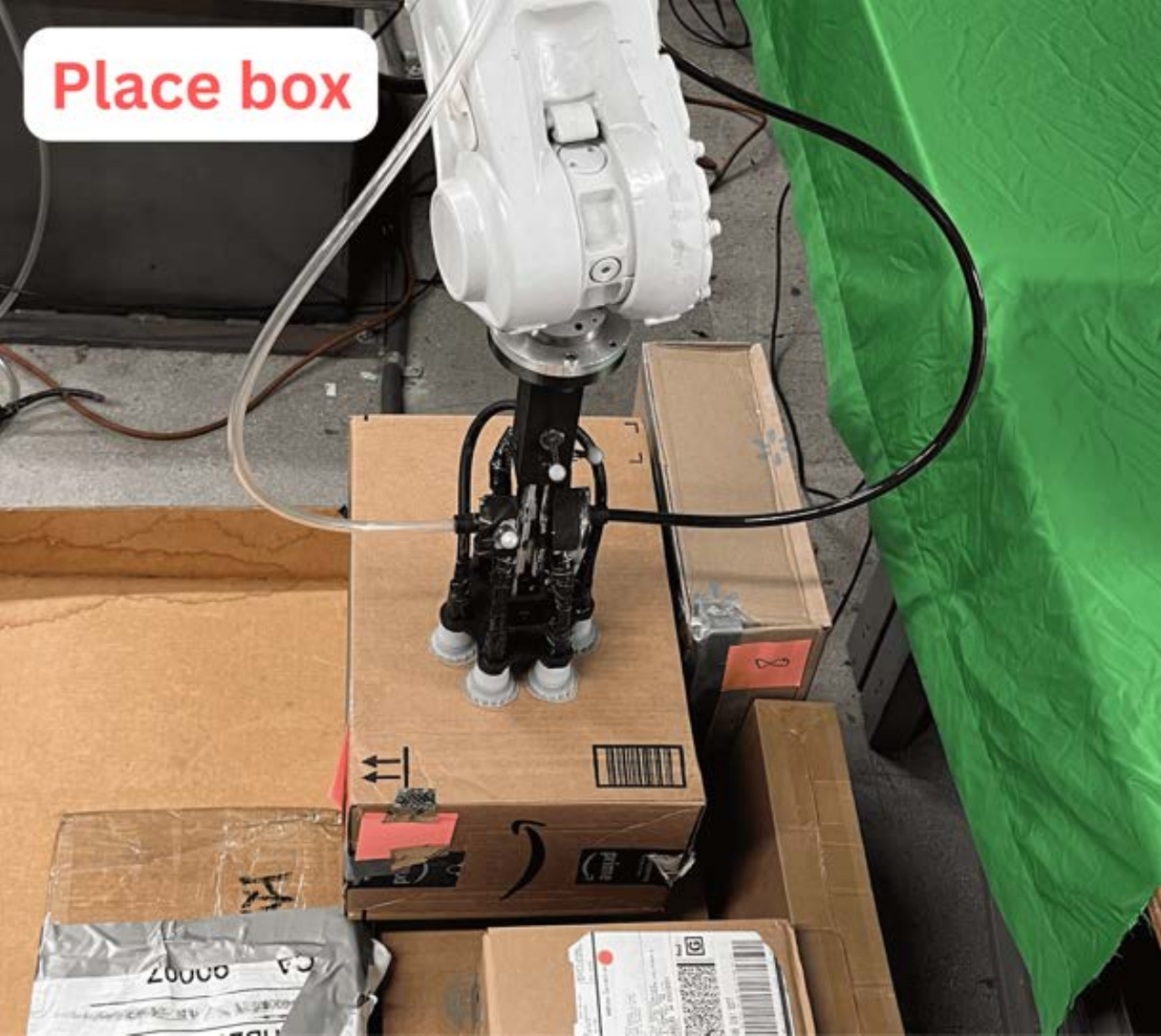}
        \vspace{-15pt}
        \caption{}
    \end{subfigure}
    \vspace{-15pt}
    \caption{The real-world experiments with an ABB robot and a suction-cup end effector. The system maintains a buffer of size 3. (a) The robot grasps the box by its front face, (b) reorients it to a transportable state and (c) places it in the bin at the predicted location.}
    \label{fig:real_experiment_figure}
    \vspace{-15pt}
\end{figure}

\subsection{Real World Experiments}
\label{sec:real_experiments}

We construct a physical packing environment to validate our method (Figure~\ref{fig:real_experiment_figure}). The setup uses an ABB IRB~2600 robot with a custom end-effector of six simultaneously actuated suction cups. The bin is \SI{92}{\centi\meter} $\times$ \SI{112}{\centi\meter} $\times$ \SI{80}{\centi\meter}, and boxes range from \SI{10}{\centi\meter} to \SI{45}{\centi\meter}. 
To simulate a replaceable buffer, a new package is manually inserted after each pick. For time-based evaluation, we measure the duration to grasp an object from a fixed home pose, and the face-specific transport times. Manual buffer replacement and the constant time for the robot to return from the bin to the home pose before the next pick action are excluded from measurements, since they are identical across all tests.



Each box surface is categorized into one of three predefined categories, as described previously, and assigned a fixed trajectory duration for time-parameterized motion planning. For smooth surfaces (Category~1), transport completes in 3\,s. Surfaces with plastic wrapping or tape (Category~2), experienced 8 out of 10 failures at 3\,s due to suction detachments, requiring a transport time of 6s to ensure stability. Category~3 required  10\,s to mitigate the highest failure frequency. Reorientations add further costs: 6\,s for the front face and 12\,s for the side. Back-face reorientations were kinematically infeasible in our setup and assigned a 100\,s to penalize selection by the time-aware policy. We assign scalar time costs to maintain focus on time-aware selection.

We evaluate the physical setup on our proposed method, STEP-3, and ReorientSpace-3, each executed across 5 runs. 
STEP-3 achieves a space utilization of $60\%$ with a time cost of $291$\,s, whereas ReorientSpace-3 reaches $63\%$ but requires $404$\,s. These results highlight that STEP-3 attains comparable packing efficiency at less operational time, demonstrating the practical benefit of time-aware selection.

\section{Conclusions}
This paper presents a preference-conditioned, Transformer-based multi-objective reinforcement learning framework for online 3D bin packing that models the trade-off between space utilization and operational time. Formulated as multi-candidate selection, it generalizes across candidate sizes and operational preferences. Although real-world complexities such as deformable objects and unstable bin dynamics are abstracted, the framework highlights the value of time-aware spatial reasoning and provides a foundation for realistic robotic packing.

\section{Acknowledgment}
This work is supported in part by Amazon Robotics. The
opinions expressed are those of the authors and do not
necessarily reflect the opinions of the sponsors.

\bibliographystyle{IEEEtran}
\bibliography{references.bib}

\end{document}